\pdfoutput=1

\documentclass[11pt,dvipsnames]{article}

\usepackage[]{acl}

\usepackage{times}
\usepackage{latexsym}

\usepackage[T1]{fontenc}

\usepackage[utf8]{inputenc}

\usepackage{microtype}

\usepackage{times}
\usepackage{latexsym}

\usepackage[T1]{fontenc}
\usepackage{makecell} 
\usepackage{multirow} 
\usepackage{multicol}
\usepackage{makecell}
\usepackage{tabularx}
\usepackage{color}
\usepackage{listings}
\usepackage{subcaption}
\usepackage{url}
\usepackage{latexsym}
\usepackage[pdftex]{graphicx}
\usepackage[inline]{enumitem}
\usepackage{array}
\usepackage{setspace}
\usepackage{framed}
\usepackage{xspace}
\usepackage{amsmath}
\usepackage{amssymb}
\usepackage{cleveref}
\usepackage{booktabs}
\usepackage{colortbl}
\usepackage{lscape}
\usepackage{adjustbox}
\usepackage{rotating}
\usepackage{arydshln}
\usepackage{soul}
\usepackage[normalem]{ulem} 
\usepackage{subcaption}
\usepackage{soul}
\usepackage{tikz}
\usetikzlibrary{calc}
\usepackage{amsmath}
\usepackage{xspace}
\usepackage{diagbox}
\usepackage{dirtytalk}
\usepackage{textgreek}

\usepackage{arydshln}
\usepackage{afterpage}

\usepackage{microtype}

\newcounter{srCounter}
\newif\ifsrvar
\srvartrue
\ifsrvar
\newcommand{\seb}[1]{{\small \color{red} \refstepcounter{srCounter}\textsf{[SR]$_{\arabic{srCounter}}$:{#1}}}}
\else
\newcommand{\seb}[1]{}
\fi

\newcounter{fpCounter}
\newif\iffpvar
\fpvartrue
\iffpvar
\newcommand{\fabio}[1]{{\small \color{blue} \refstepcounter{fpCounter}\textsf{[FP]$_{\arabic{fpCounter}}$:{#1}}}}
\else
\newcommand{\fabio}[1]{}
\fi

\newcounter{apCounter}
\newif\ifapvar
\apvartrue
\ifapvar
\newcommand{\piktus}[1]{{\small \color{orange} \refstepcounter{apCounter}\textsf{[AP]$_{\arabic{apCounter}}$:{#1}}}}
\else
\newcommand{\piktus}[1]{}
\fi

\newcounter{vkCounter}
\newif\ifvkvar
\vkvartrue
\ifvkvar
\newcommand{\vladk}[1]{{\small \color{green} \refstepcounter{vkCounter}\textsf{[VK]$_{\arabic{vkCounter}}$:{#1}}}}
\else
\newcommand{\vladk}[1]{}
\fi

\newcounter{egCounter}
\newif\ifegvar
\egvartrue
\ifegvar
\newcommand{\egrave}[1]{{\small \color{purple} \refstepcounter{egCounter}\textsf{[EG]$_{\arabic{egCounter}}$:{#1}}}}
\else
\newcommand{\egrave}[1]{}
\fi

\newcounter{syCounter}
\newif\ifegvar
\egvartrue
\ifegvar
\newcommand{\scott}[1]{{\small \color{violet} \refstepcounter{syCounter}\textsf{[SY]$_{\arabic{syCounter}}$:{#1}}}}
\else
\newcommand{\scott}[1]{}
\fi

\newcounter{sbCounter}
\newif\ifegvar
\egvartrue
\ifegvar
\newcommand{\samuel}[1]{{\small \color{cyan} \refstepcounter{sbCounter}\textsf{[SB]$_{\arabic{sbCounter}}$:{#1}}}}
\else
\newcommand{\samuel}[1]{}
\fi

\newcounter{yzCounter}
\newif\ifegvar
\egvartrue
\ifegvar
\newcommand{\yizhong}[1]{{\small \color{Aquamarine} \refstepcounter{yzCounter}\textsf{[YZ]$_{\arabic{yzCounter}}$:{#1}}}}
\else
\newcommand{\yizhong}[1]{}
\fi

\newcounter{msCounter}
\newif\ifegvar
\egvartrue
\ifegvar
\newcommand{\minjoon}[1]{{\small \color{purple} \refstepcounter{msCounter}\textsf{[MS]$_{\arabic{msCounter}}$:{#1}}}}
\else
\newcommand{\minjoon}[1]{}
\fi

\newcommand{\fever}{FEV\xspace}
\newcommand{\trex}{T-REx\xspace}
\newcommand{\zsre}{zsRE\xspace}
\newcommand{\nq}{NQ\xspace}
\newcommand{\hotpot}{HoPo\xspace}
\newcommand{\trivia}{TQA\xspace}
\newcommand{\eli}{ELI5\xspace}
\newcommand{\wow}{WoW\xspace}
\newcommand{\tqa}{TQA\xspace}

\newcommand{\sphere}{\textsc{Sphere}\xspace}

\newcommand{\ccnet}{CCNet\xspace}

\newcommand{\aic}{\emph{AIC}\xspace}
\newcommand{\aeic}{\emph{AEIC}\xspace}

\newcommand{\fid}{\textsc{FiD}\xspace}
\newcommand{\dprmulti}{\textsc{DPR\textsubscript{multi}}\xspace}
\newcommand{\dprweb}{\textsc{DPR\textsubscript{web}}\xspace}

\newcommand{\fidmulti}{\fid{+}\dprmulti\xspace}
\newcommand{\fidweb}{\fid{+}\dprweb\xspace}
\newcommand{\fidbm}{\fid{+}\textsc{BM25}\xspace}

\newcolumntype{R}[1]{>{\raggedleft\let\newline\\\arraybackslash\hspace{0pt}}m{#1}}
\newcolumntype{C}[1]{>{\centering\let\newline\\\arraybackslash\hspace{0pt}}m{#1}}

\title{The Web Is Your Oyster - Knowledge-Intensive NLP \\ against a Very Large Web Corpus}

\newcommand{\fair}{$^1$}
\newcommand{\ucl}{$^2$}
\newcommand{\uman}{$^3$}
\newcommand{\ens}{$^4$}
\newcommand{\inria}{$^5$}
\newcommand{\uw}{$^6$}
\newcommand{\uclfair}{$^{1,2}$}
\newcommand{\umanfair}{$^{1,3}$}
\newcommand{\fairensinria}{$^{1,4,5}$}

\author{
Aleksandra Piktus\fair{}
Fabio Petroni\fair{}
Yizhong Wang\uw{}
Vladimir Karpukhin\fair{}
Dmytro Okhonko\fair{} \\
\textbf{Samuel Broscheit\umanfair{} Gautier Izacard\fairensinria{} Patrick Lewis\uclfair{} Barlas Oğuz\fair{}} \textbf{Minjoon Seo}\fair{} \\
\textbf{Edouard Grave\fair{} Wen-tau Yih\fair{} Sebastian Riedel\uclfair{}} \\
\\
\fair{}Facebook AI Research \ \ucl{}University College London \\ 
\uman{}University of Mannheim \ \ens{}ENS, PSL University \ \inria{}Inria \ \uw{}University of Washington \\
{\tt piktus@fb.com} \\
}

\date{}

\begin{document}
\maketitle

\begin{abstract}
In order to address increasing demands of real-world applications, the research for knowledge-intensive NLP (KI-NLP) should advance by capturing the challenges of a \emph{truly} open-domain environment: web-scale knowledge, lack of structure, inconsistent quality and noise. To this end, we propose a new setup for evaluating existing knowledge intensive tasks in which we generalize the background corpus to a universal web snapshot.
We investigate a slate of NLP tasks which rely on knowledge - either factual or common sense, and ask systems to use a subset of \ccnet---the \sphere corpus---as a knowledge source. In contrast to Wikipedia, otherwise a common background corpus in KI-NLP, \sphere is orders of magnitude larger and better reflects the full diversity of knowledge on the web.
Despite potential gaps in coverage, challenges of scale, lack of structure and lower quality, we find that retrieval from \sphere enables a state of the art system to match and even outperform Wikipedia-based models on several tasks. 
We also observe that while a dense index can outperform a sparse BM25 baseline on Wikipedia, on \sphere this is not yet possible. To facilitate further research and minimise the community's reliance on proprietary, black-box search engines, we share our indices, evaluation metrics and infrastructure.
\end{abstract}

\section{Introduction}
\begin{table}[ht]
  \renewcommand{\arraystretch}{1.2}
  \centering
  \resizebox{\linewidth}{!}{    
  \fontsize{8.6}{10.1}\selectfont
  \begin{tabular}{>{\raggedleft\arraybackslash}p{.1\textwidth}p{.35\textwidth}} \toprule
      \textbf{\textsc{Query}} & Who is Joëlle Sambi Nzeba? \\
      \midrule
      \textbf{\textsc{Wikipedia}} & \textcolor{red}{No results found for} \textcolor{blue}{Joëlle Sambi Nzeba}\textcolor{red}{.} \\
      \textbf{\sphere} & 
             {\fontfamily{cmss}\selectfont{[...] Joëlle Sambi. She was born in Belgium and grew up partly in Kinshasa (Congo). She currently lives in Brussels. She is a writer and slammer, alongside her activism in a feminist movement. She is an award-winning author of fiction with Le Monde est gueule de chèvre (novel, 2007) and Je ne sais pas rêver (short-stories, 2002). Joëlle Sambi questions situations of powerlessness in social matters and raises questions about identity, [...]}} \\ 
             \textbf{URL} & \url{https://www.buala.org/en/mukanda/musala-worf} \\
             \bottomrule
    \end{tabular}
  }
  \caption{Web covers more knowledge than Wikipedia. We pose a question about an activist listed in the \emph{\href{https://en.wikipedia.org/wiki/Wikipedia:WikiProject\_Women\_in\_Red/Feminists?oldid=1032819455}{Women in Red}} project - an initiative mobilizing the community to fill the Wikipedia gender gap, and the top retrieval result from \sphere. At the time of writing, Joëlle Sambi Nzeba does not have a Wikipedia page.
  }
  \label{tab:example_women}
\end{table}

\begin{table*}[t!]
    \centering
\resizebox{\linewidth}{!}{    
    \fontsize{8.4}{10.1}\selectfont \setlength{\tabcolsep}{0.5em}
    \begin{tabular}{cccccc}
        \toprule
        Name & Reference & Corpus source & Task & \#passages & \#documents \\
        \midrule
        KILT &  \citet{petroni-etal-2021-kilt} & Wikipedia snapshot  & Multitask & 22M & 5.9M \\
        \midrule
        TriviaQA & \citet{joshi2017triviaqa} & Bing search results & ODQA & - & 662K \\ 
        MSMarco & \citet{bajaj2018ms} & Bing search results & ODQA & 8.8M &  3.2M \\
        ComplexWQ & \citet{talmor-berant-2018-web} & Web search snippets & ODQA & 12.7M & - \\
        Eli5 & \citet{fan2019eli5} & Common Crawl search results & ODQA & - & 27.2M \\
        Internet Augmented Dialog & \citet{komeili2021internetaugmented} & \ccnet snapshot & Dialog & 250M &  109M \\
        \sphere & \textit{ours} & \ccnet snapshot  & Multitask & 906M &  134M \\
        \bottomrule
    \end{tabular}
    }
\caption{Sizes of large scale unstructured web corpora for KI-NLP tasks.
}
\label{tab:corpus_size}
\end{table*}

\begin{table}[t!]
  \centering
  \resizebox{\columnwidth}{!}{    
    \fontsize{8.4}{10.1}\selectfont \setlength{\tabcolsep}{0.5em}
    \begin{tabular}{rcc}
    
      \toprule
      Shortcut & Dataset & \multicolumn{1}{c}{Reference} \\
      \midrule
      \multicolumn{3}{c}{\textbf{KILT}} \\ 
      \midrule

      \textbf{FEV} &FEVER & \citet{Thorne18Fever} \\ 
      \textbf{T-REx} & T-REx & \citet{elsahar2019t} \\ 
      \textbf{zsRE} & Zero Shot RE & \citet{levy2017zero} \\ 
      \textbf{NQ} & Natural Questions & \citet{kwiatkowski-etal-2019-natural} \\ 
      \textbf{HoPo} & HotpotQA & \citet{yang2018hotpotqa} \\ 
      \textbf{TQA} & TriviaQA & \citet{joshi2017triviaqa} \\ 
      \textbf{ELI5} & ELI5 & \citet{fan2019eli5} \\ 
      \textbf{WoW} & Wizard of Wikipedia & \citet{dinan2018wizard} \\  
      
      \midrule
      \multicolumn{3}{c}{\textbf{Common Sense}} \\ 
      \midrule
      
      \textbf{COPA} & COPA & \citet{roemmele2011copa} \\ 
      \textbf{PIQA} & PIQA & \citet{bisk2020piqa} \\ 
      \textbf{H-SWAG} & HellaSWAG & \citet{zellers2019hellaswag} \\ 
      \textbf{CSQA} & CommonsenseQA & \citet{talmor2019commonsenseqa} \\ 
      \textbf{\textalpha NLI} & \textalpha NLI & \citet{bhagavatula2020abductive} \\ 
      \textbf{NumS} & NumerSense & \citet{lin2020birds} \\ 
      \textbf{WG} & WinoGrande & \citet{sakaguchi2020winogrande} \\ 
      \textbf{SocIQa} & SocialIQa & \citet{sap2019social} \\  
      \textbf{CosQA} & CosmosQA & \citet{huang2019cosmos} \\
    
      \bottomrule
    \end{tabular}
  }
  \caption{Downstream tasks we consider.
  }
  \label{tab:datasets}
\end{table}

The ability to access and manipulate knowledge has become one of the core features of modern NLP systems. Knowledge-intensive NLP (KI-NLP) tasks such as fact checking, open-domain question answering~(ODQA) and entity linking typically specify a source of factual knowledge necessary to provide an explainable solution.
Often, this source is Wikipedia~\cite{dinan2018wizard,fever-2020-fact, kwiatkowski-etal-2019-natural}, for obvious reasons: it tends to be highly accurate, it is well-structured and small enough to test computationally demanding architectures. Still, there exist many reasons to look beyond Wikipedia. First, it covers a lot of ground, but certainly not everything, and in practice many information needs cannot be fulfilled based on Wikipedia alone~\cite{redi2021taxonomy}. Second, even for topics it does cover, there might be biases that cannot be resolved without looking at a broader context~\cite{ DBLP:journals/corr/WagnerGG16, DBLP:journals/corr/Graells-Garrido15}.
Unsurprisingly, Wikipedia has also never gained much traction in common sense NLP. Contrary to the factual knowledge, common sense is believed to be universally accepted by humans while stated more implicitly \cite{xie2021commonsense}. There have been many attempts to capture such implicit knowledge vie common sense knowledge bases \cite{speer2012conceptnet, sap2019atomic, sumithra2020genericskb}. While yielding great results, their supervised nature makes them hard to generalize and expand.

By virtue of its sheer scale, the web promises access to knowledge both broader and more in-depth than Wikipedia, providing not only sheer facts but also context useful in inferring rules of common sense reasoning. Along with the benefits, however, come new challenges---lack of structure, inconsistent document quality and noisy or harmful content on one hand \cite{luccioni-viviano-2021-whats}, increasing infrastructural demands on the other. Today, the impact of these challenges on knowledge tasks is not clear---while work investigating the use of web in KI-NLP exists, it usually relies on commercial, black-box search engines, focuses on individual tasks, primarily ODQA~\cite{joshi2017triviaqa, bajaj2018ms,talmor-berant-2019-multiqa,DBLP:journals/corr/abs-2112-09332}, or only uses general web content at pre-training time~\cite{guu2020realm, borgeaud2022improving, lewis2020pretraining}.

We propose to use a web corpus as a universal, uncurated and unstructured knowledge source for multiple KI-NLP tasks at once. 
We aim to answer the following question: \emph{what impact does replacing Wikipedia with a large-scale web corpus have on the performance of knowledge-intensive systems?} Specifically, should we expect them to improve, since for a given fact, there is more potential evidence on the web, or degrade, due to uncurated nature of the data? Multiple factors such as the scope of knowledge covered by the corpus, the ability of retrievers to generalize across downstream tasks and the scalability of the solution may contribute to the answer. We propose a unified retrieval infrastructure and analyze these aspects of our setup in depth. We then explore if our web index can serve as a knowledge source in common sense NLP.

We leverage an open web corpus coupled with strong retrieval baselines instead of a black-box, commercial search engine---an approach which facilitates transparent and reproducible research and opens up a path for future studies comparing search engines optimised for humans with retrieval solutions designed for neural networks.
We use a subset of \ccnet~\cite{wenzek-etal-2020-ccnet} covering 134M documents split into 906M passages as the web corpus which we call \sphere. While far from the full web scale, \sphere is orders of magnitude larger than previously studied knowledge sources (cf. Table~\ref{tab:corpus_size}).
We consider two retrieval architectures---BM25~\cite{robertson2009probabilistic} and DPR~\cite{karpukhin-etal-2020-dense}, and combine them with a \fid reader component~\cite{izacard2020leveraging}. To facilitate large-scale dense retrieval, we open-source \emph{distributed-faiss}---a wrapper around the FAISS similarity search library~\cite{JDH17}, simplifying the distribution of indices across machines. We use KILT~\cite{petroni-etal-2021-kilt}, a standard KI-NLP benchmark, as well as a range of common sense tasks listed in Table~\ref{tab:datasets} to evaluate our work.

Despite inconsistent quality of the web and the fact that KILT was specifically designed to query knowledge from Wikipedia, we find that \sphere-based models can match or outperform baselines grounded in Wikipedia on a subset of KILT tasks. In some cases this holds even when we aggressively filter \sphere by removing not just Wikipedia itself but also content that looks like it. Moreover, we find that retrieval from \sphere can improve the common sense abilities of \fid, despite the lack of \emph{explicit} common sense knowledge in retrieved passages. To our knowledge, this is the first time a general purpose search index improves language models on common sense tasks.

We also find ample room for future work: while dense retrieval outperforms sparse methods in most prior work, in our case the opposite is true. How to develop large scale and \emph{universal} dense indices supporting a multitude of tasks hence remains an open question. To summarize, we make the following contributions.
\begin{itemize}
\item We replace Wikipedia with \sphere as the knowledge source for a selection of KILT tasks, achieving state of the art results on two.
\item We carry out an in-depth analysis of our web corpus and retrievers, identifying potential reasons for both gains and losses in end-to-end performance on respective tasks.
\item We show that general purpose web retrieval from \sphere can improve common sense reasoning on 8 tasks when compared to a comparable, fully parametric language model.
\item We release sparse and dense indices of \sphere and open-source \emph{distributed-faiss}.
\end{itemize}
\begin{figure*}[ht]
\centering
\includegraphics[width=\textwidth]{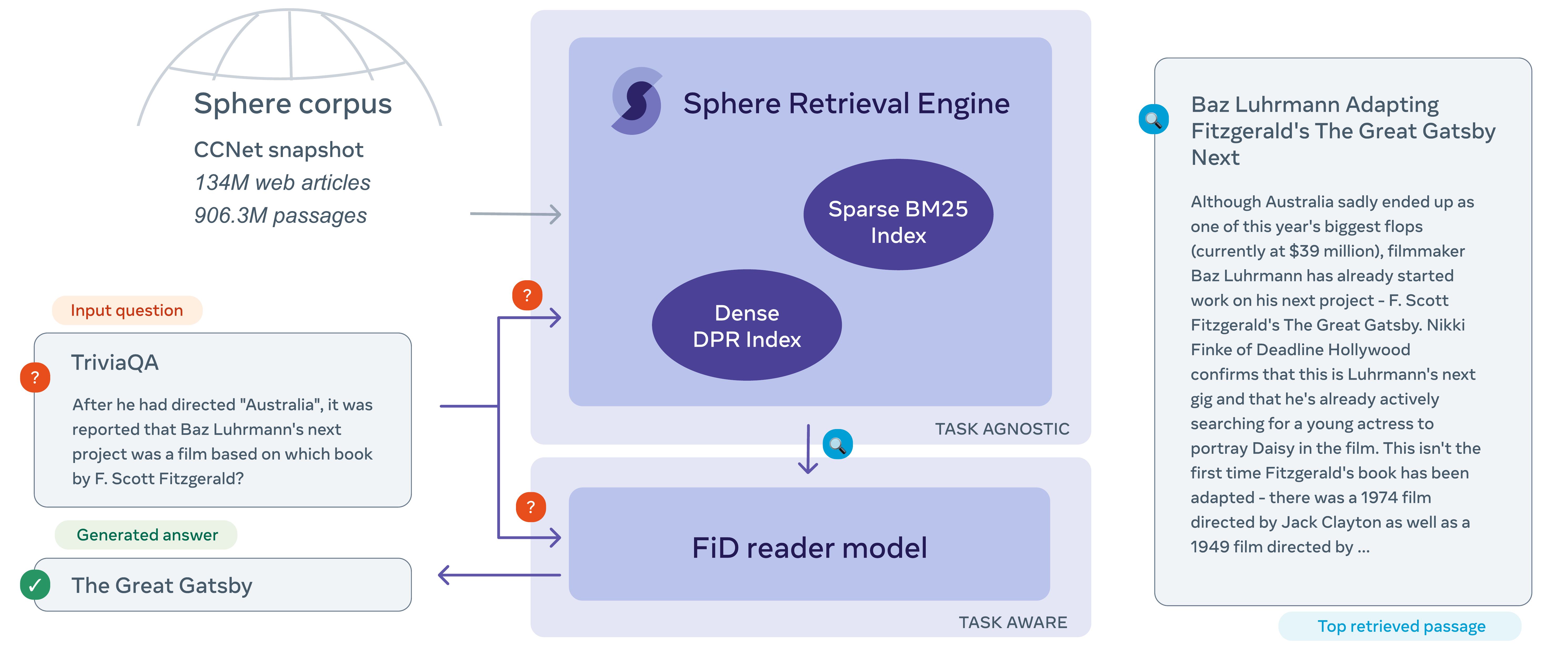}  
\caption{An outline of an end to end system. We build a universal, task-independent index of the \sphere corpus offline. We experiment with two retrieval architectures - BM25 and DPR. We then train a \fid reader model with passages retrieved from \sphere for each downstream tasks separately.}
\label{fig:splash}
\end{figure*}

\section{Background}
\subsection{Knowledge-intensive NLP Tasks}
We typically call an NLP task \emph{knowledge-intensive} if a human would not be reasonably expected to solve it without access to an external knowledge source. KI-NLP tasks are usually solved with retriever-reader systems: first, a retriever surfaces a small set of relevant documents from the knowledge source, then a reader uses the context to generate an answer~\cite{DBLP:journals/corr/ChenFWB17,lewis2020retrievalaugmented,guu2020realm}. For the purpose of this work we expand our definition of knowledge-intensive to any NLP task which, beyond core language capabilities, also requires knowledge---be it factual or common sense---to solve the problem at hand.

\subsection{Retrieval models}
\label{ssub:retrievalbg}
We consider two retrieval architectures. BM25 \cite{robertson2009probabilistic} is a popular \emph{sparse} model, where queries and documents are represented as high-dimensional, \emph{sparse} vectors, with dimensions corresponding to vocabulary terms and weights indicating their importance. DPR~\cite{karpukhin-etal-2020-dense} is a \emph{dense} model which embeds queries and documents into a latent, real-valued vector space of a much lower dimensionality---an idea originating from the Latent Semantic Analysis~\cite{deerwester1990indexing}. DPR is based on a neural bi-encoder architecture with passages and queries embedded with separate text encoders. 
Although both sparse and dense models use the distance in the vector space as the relevance function, they need different indexing schemes to support efficient retrieval.

\subsection{Reader models}
Reader models typically consume a set of context documents retrieved from a knowledge source and the task input, and return the output---either a class label or text.
In this work, we use an abstractive Fusion-in-Decoder (\fid) reader from \citet{izacard2020leveraging} --- an encoder-decoder architecture, where each context document is concatenated with the input and embedded by the encoder. In the decoder, attention is performed over encoded passages and then the output is generated.
\section{Search Infrastructure}
A question equally important to the choice of the knowledge corpus itself pertains to the feasibility of implementing a research-friendly search infrastructure on top of it.
An index is a data structure which stores representations of corpus documents, built with the objective of optimizing the retrieval efficiency. For sparse methods, this goal is typically achieved with an inverted index---a space-efficient technique entertaining the support of multiple robust libraries such as Pyserini~\cite{lin2021pyserini}. Efficient dense retrieval is enabled by maximum inner product search algorithms \cite{shrivastava2014asymmetric, guo2015quantization} leveraged by tools like FAISS \cite{JDH17}, a robust library for similarity search and clustering of dense vectors.
As the size of the text corpus grows, a FAISS index may exceed typical, single-server hardware limits for both GPU and RAM. Two main approaches for handling scale emerge: compression of the document embeddings and distribution of the index over multiple servers. Good compression rates can be achieved with quantizers available in FAISS out-of-the-box or with more sophisticated bi-encoder training pipelines~\cite{yamada2021hash, zhan2021quant}---this may help reduce the index size by a few times factor but does not solve the core scaling issue. The ability to distribute a FAISS index is what we address with our open-source release of \emph{distributed-faiss}\footnote{
\textit{ \url{https://github.com/facebookresearch/distributed-faiss}}}
At indexing time, the \emph{distributed-faiss} client receives batches of embeddings to be indexed and routes them to the index servers guaranteeing a balanced data distribution. At retrieval time, the client queries all servers and aggregates the results. The service is model-independent and operates with supplied embeddings and metadata.
We also release indices of \sphere
\footnote{
\textit{\url{https://github.com/facebookresearch/Sphere}} }
both for the sparse retrieval baseline, compatible with Pyserini, and our best dense model compatible with \textit{distributed-faiss}.

\section{Experimental setup}
\paragraph{Background Corpora.}
We experiment with two universal (non task-specific) knowledge sources. First, the KILT knowledge source based on the 2019/08/01 Wikipedia snapshot, comprising 5.9M articles split into 22.2M passages of 100 tokens. We refer to this corpus as Wikipedia in the reminder of this paper. Second, we use \ccnet~\cite{wenzek-etal-2020-ccnet} to create our web corpus. \ccnet processes Common Crawl by performing deduplication, language identification and quality filtering (articles are split into three quality tiers: \textit{head}, \textit{middle} and \textit{tail} based on perplexity under a Wikipedia-based language model). We use the \textit{head} tier of a single \ccnet snapshot in English, additionally excluding all articles containing \url{wikipedia.org} in their URL. We pick the \ccnet snapshot corresponding to the August 2019 Common Crawl snapshot\footnote{\url{https://commoncrawl.org/2019/08/august-2019-crawl-archive-now-available/}} as it is temporally the closest to the KILT knowledge source. It consists of 134M web articles and yields 906.3M passages of 100 tokens. We call this corpus \sphere in the remainder of this paper. 

\paragraph{Downstream Tasks.}
We use the KILT benchmark as a KI-NLP evaluation suite for our work. KILT contains 11 tasks, split into 5 categories: fact checking, entity linking, slot filling, ODQA and dialog. Since entity linking is intrinsically tied to the underlying Wikipedia corpus, as entity labels are equivalent to the titles of Wikipedia pages representing respective entities, we choose to exclude it from our work. We also experiment with a collection of common sense tasks---see Table~\ref{tab:datasets} for the full list of datasets and shortcuts we use to refer to them. For each downstream tasks, we train a \fid reader model, using T5-base \cite{raffel2019exploring} as initialization for the KILT tasks, and T5-large for the common sense tasks. Otherwise, we follow the finetuning setup proposed by \fid authors. Unless otherwise stated, we train \fid with the top 100 passages retrieved from considered corpora. 

\paragraph{Baseline retrievers.}
We experiment with three retrieval baselines: BM25, \dprmulti, a variant of DPR pre-trained in a multi-task fashion on KILT by~\citet{maillard-etal-2021-multi}, and \dprweb, a new DPR model trained for the purpose of this work (details in the next paragraph). Both DPR models use 768-dim encoders.
We use Pyserini~\cite{lin2021pyserini} to build the BM25 index of \sphere and Wikipedia and \emph{distributed-faiss} to build the dense indices. We build an HNSWSQ8 FAISS index of \sphere, with 16 physical and 32 logical nodes providing a good accuracy and latency of 100--200 ms. The index, which consists of the embeddings and metadata occupies 1.9TB of disk space. We use a flat FAISS index for Wikipedia. We follow \citet{karpukhin-etal-2020-dense} and index 100-token long passages along with article titles.
We apply all three retrievers to \sphere. As DPR has been repeatedly shown to outperform BM25 in retrieval from Wikipedia on KI-NLP tasks~\cite{karpukhin-etal-2020-dense,maillard-etal-2021-multi}, we skip BM25 in those experiments. Due to the lack of retrieval supervision, we only consider BM25 in our common sense experiments.

\paragraph{Training \dprweb.}
Our goal is to leverage \sphere in the training of a DPR web retriever. Given the lack of any explicit retrieval supervision over \sphere, we use two proxy metrics to track performance: \emph{answer-in-context@k} (\aic{@k}), indicating the fraction of examples for which there exists a passage containing the gold answer among the top-$k$ retrieved ones; and \emph{answer+entity-in-context@k} (\aeic{@k}), indicating the fraction of examples for which among the top-$k$ passages there exists one containing both the gold answer and the main entity of the datapoint---we use the Wikipedia title of the gold retrieval passage defined by KILT as the main entity.
We train \dprweb by finetuning a PAQ-based \cite{10.1162/tacl_a_00415} bi-encoder checkpoint \cite{oguz2021domainmatched} for 40 epochs on 16 GPUs. We source finetuning data from KILT tasks compatible with the \aic metric---so those with short-form textual answers (\trex, \zsre, \nq, \hotpot and \trivia) and apply the model in zero-shot fashion to the remaining tasks.
We balance the number of datapoints per dataset by sampling with the same rates as in \citet{maillard-etal-2021-multi}.
For each datapoint we source context passages in 4 ways (with in-batch negatives in all cases):
(1) gold Wikipedia passages as positives and BM25-based negatives (the same as in original DPR);
(2) gold Wikipedia passages as positives and hard Wikipedia negatives (BM25 index based, the same as were used to finetune \dprmulti);
(3) weakly supervised web positives from the BM25 \sphere index;
(4) weakly supervised web positives from the \dprmulti \sphere index.
We obtain the weakly supervised positive web passages by picking the top result returned by respective baseline retrievers containing the gold answer of a given datapoint. We use the batch size 32 and default DPR hyperparameters otherwise.
\begin{table*}[t!]
\centering
\resizebox{0.9\linewidth}{!}{
  \setlength{\tabcolsep}{0.5em}
  \begin{tabular}{ccccccccc}
    \toprule
    & Fact Check.  & \multicolumn{2}{c}{Slot Filling} & \multicolumn{4}{c}{Open Domain QA} & \multicolumn{1}{c}{Dial.}  \\
    \cmidrule(lr){2-2} \cmidrule(lr){3-4} \cmidrule(lr){5-8} \cmidrule(lr){9-9}
    Model  & \textbf{\fever}  & \textbf{\trex} & \textbf{\zsre}  & \textbf{\nq} & \textbf{\hotpot} & \textbf{\trivia} & \textbf{\eli}  & \textbf{\wow}   \\
    \cmidrule(lr){2-4} \cmidrule(lr){5-7} \cmidrule(lr){8-8} \cmidrule(lr){9-9} 
    & \multicolumn{3}{c}{Accuracy} & \multicolumn{3}{c}{Exact Match} & \multicolumn{1}{c}{RL} & \multicolumn{1}{c}{F1} \\
    \midrule
    
    \multicolumn{9}{c}{\textbf{Wikipedia}} \\
    \midrule
    \cite{glass2021robust}  & - & \textbf{84.36} & \underline{72.55} & - & - & - & - & - \\
    \cite{petroni-etal-2021-kilt} \begin{tiny} BART+DPR \end{tiny} & \underline{86.74} & 59.16 & 30.43 & 41.27 & 25.18 & 58.55 & 17.41 & 15.19 \\
    \cite{petroni-etal-2021-kilt}  \begin{tiny} RAG \end{tiny} & 86.31 & 59.2 & 44.74 & \underline{44.39} & 26.97 & \underline{71.27} & 14.05 & 13.11\\
    \cite{maillard-etal-2021-multi}  & 86.32 & -  & 57.95 & 39.75 & \underline{31.77} & 59.60 & -  & 15.33 \\
    \cite{krishna-etal-2021-hurdles}  &- &- &- &- &- & -& \textbf{23.4} & -\\
    \cite{paranjape2021hindsight}  & -& -& -& -&- &- &- & \textbf{19.19} \\
    \midrule
    \fidmulti & 88.99 & 82.19 & 71.53 &  49.86 & 36.90 & 71.04 & 16.45 & 15.66 \\
    \fidweb & 89.03 & 81.34 &  \textbf{73.96} & \textbf{51.59} &  \textbf{38.27} & 72.73 & 15.91 & 15.45 \\
    \midrule
    \multicolumn{9}{c}{\textbf{\sphere}} \\
    \midrule
    \fidmulti & 85.74 & 52.06 & 28.47 & 45.15 & 27.29 & 67.49 & 16.14 & 15.22 \\
    \fidweb & 87.43 & 57.02 & 36.55 & 48.61 & 31.64 & 73.06 & 15.76 & 15.29 \\
    \fidbm & \textbf{89.12} & 62.12 & 43.92 & 46.05 & 34.10 &  \textbf{78.21} & 15.59 & 17.28 \\
    \bottomrule
  \end{tabular}
}
\caption{Downstream evaluation results on the test set as per KILT leaderboard. We present results for published baselines (top section), our Wikipedia-based models (middle section) and \sphere-based models (bottom section). SOTAs in bold, previous SOTAs underlined.}
\label{tab:reader_test}
\end{table*}
\begin{table}[t!]
\centering
\resizebox{\columnwidth}{!}{    
    \fontsize{8.4}{10.1}\selectfont \setlength{\tabcolsep}{0.5em}
\begin{tabular}{cccccccc}

\toprule
  \textbf{\fever}  & \textbf{\trex} & \textbf{\zsre}  & \textbf{\nq} & \textbf{\hotpot} & \textbf{\trivia} & \textbf{\eli}  & \textbf{\wow}   \\

\midrule
\multicolumn{8}{c}{Entity in input} \\
\midrule
83.15 & 71.42 & 99.09 & 36.20 & 66.79 & 67.29 & 11.55 & 66.99 \\
\bottomrule
\end{tabular}
}
\caption{The percentage of datapoints in the dev set, for which the input contains the title of the main entity.}
\label{tab:entity_in_input}
\end{table}

\section{Results}
\subsection{KILT on \sphere}
We present our main results in Table~\ref{tab:reader_test}. It is important to remember that the KILT benchmark was designed with a specific Wikipedia snapshot in mind and examples for which no evidence was found were removed. Thus, there is a strong bias towards Wikipedia as the knowledge source, and the performance of systems using it can be considered topline. We also note that ours is the first paper to report \fid results on KILT---our baseline \fidmulti model outperforms similar DPR-based architectures across the board. In order to factor out the impact of moving to a stronger reader, we mainly focus on comparing our \sphere-based models to our Wikipedia baselines, with \fid reader in both cases. Our \sphere-based \fidbm architecture establishes a new state of the art (SOTA)\footnote{We compare to current (Nov. 2021) leaderboard results at \url{kiltbenchmark.com} published on \url{arxiv.org}.} on \fever and \trivia (see Table~\ref{tab:examples} in Appending for examples). We also note that a \fidweb model beats SOTA on \zsre, \nq and \hotpot with Wikipedia as the knowledge source.

\paragraph{Finetuning DPR.}
With \sphere as the knowledge source, \dprweb outperforms \dprmulti on all KILT tasks but \eli (see Table~\ref{tab:ans_ent} in the Appendix for retrieval evaluation), yielding notable gains downstream: +8 points on \zsre, +6 on \trivia, +5 on \trex.
Interestingly, when used to retrieve from Wikipedia, \dprweb also helps. We see gains on all tasks used in DPR finetuning except \trex, with new SOTAs on \zsre, \nq and \hotpot. By contrast, results on tasks excluded from DPR finetuning are not consistent. Unlike with \fever, where \dprweb yields SOTA both against Wikipedia and \sphere, we don't observe downstream gains for the long-form QA and dialog, which highlights the challenge of zero-shot transfer in dense retrieval.

\begin{figure*}
%
%
\begin{subfigure}{.49\textwidth}
 \includegraphics[width=\textwidth]{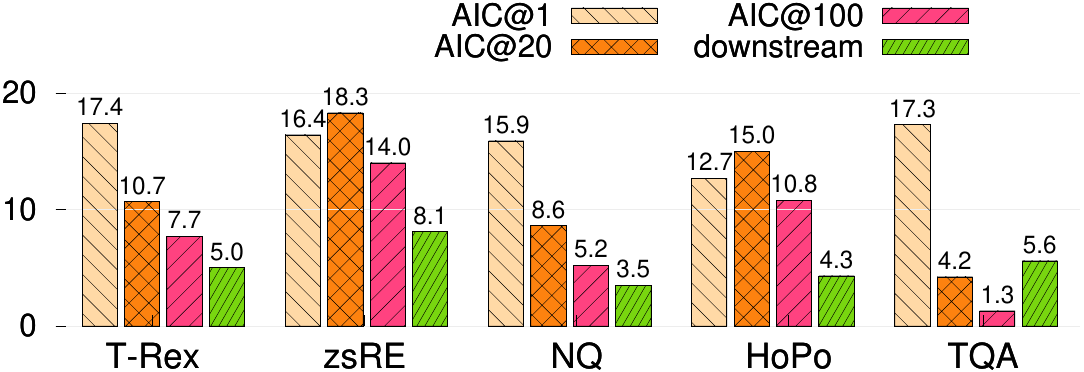}
 \caption{\dprweb vs \dprmulti}
 \label{fig:ans_ent_a}
\end{subfigure}\hspace{.01\textwidth}
\begin{subfigure}{.49\textwidth}
 \includegraphics[width=\textwidth]{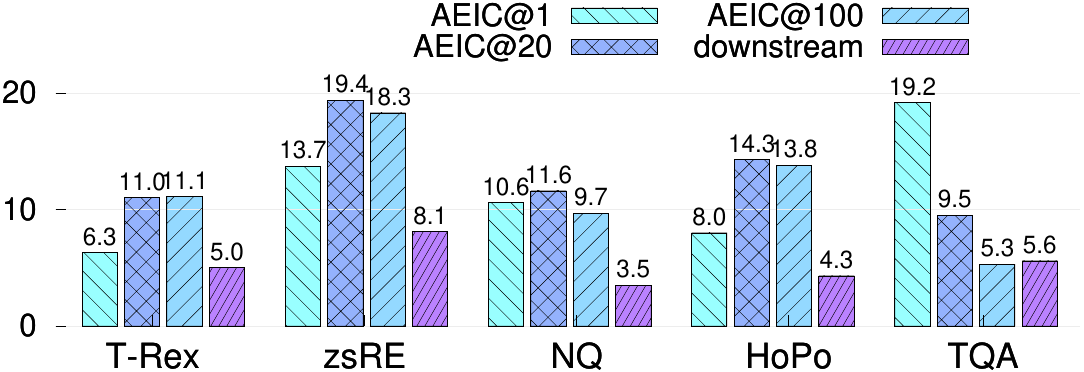}
 \caption{\dprweb vs \dprmulti}
 \label{fig:ans_ent_b}
\end{subfigure}
\vspace{1em}

\begin{subfigure}{.49\textwidth}
 \includegraphics[width=\textwidth]{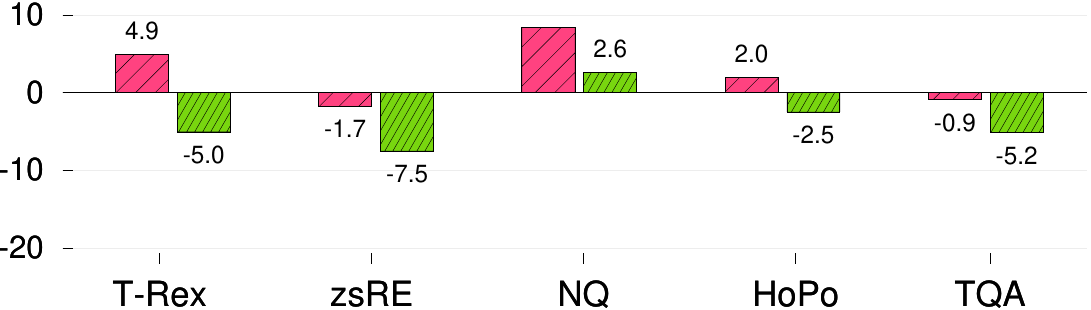}
 \caption{\dprweb vs BM25}
 \label{fig:ans_ent_c}
\end{subfigure}\hspace{.01\textwidth}
\begin{subfigure}{.49\textwidth}
 \includegraphics[width=\textwidth]{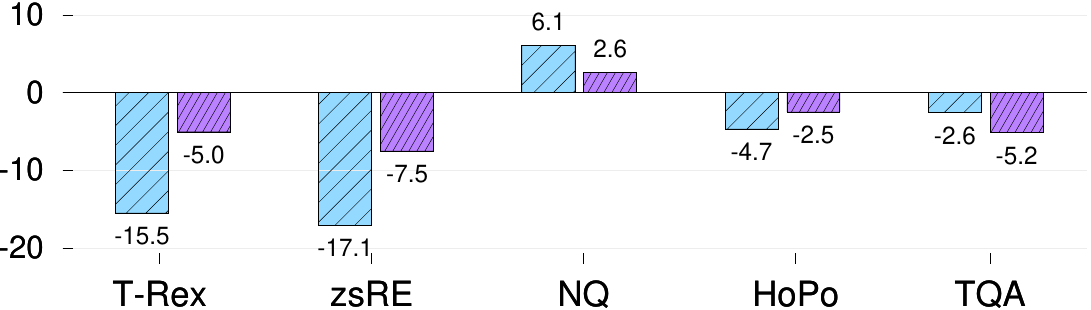}
 \caption{\dprweb vs BM25}
 \label{fig:ans_ent_d}
\end{subfigure}

\caption{The absolute change in retrieval and downstream performance between baseline retrievers and \dprweb.}
\label{fig:answer_in_context}
\end{figure*}

\begin{table*}[t!]
\centering
\resizebox{0.75\linewidth}{!}{    
  \setlength{\tabcolsep}{0.5em}
\begin{tabular}{ccccccccc}
 \toprule
\multirow{2}{*}[-0.3em]{Model} & \textbf{\fever}  & \textbf{\trex} & \textbf{\zsre}  & \textbf{\nq} & \textbf{\hotpot} & \textbf{\trivia} & \textbf{\eli}  & \textbf{\wow}   \\

 \cmidrule(lr){2-4} \cmidrule(lr){5-7} \cmidrule(lr){8-8} \cmidrule(lr){9-9} 
 & \multicolumn{3}{c}{Accuracy} & \multicolumn{3}{c}{Exact Match} & \multicolumn{1}{c}{RL} & \multicolumn{1}{c}{F1} \\
\midrule
 
Wikipedia (\fidweb) &90.93 & 80.94 & 72.39 & 54.98 & 38.04 & 71.43 & 17.88 & 16.11 \\
\sphere (\fidbm)  & 90.71 & 59.66 & 38.61 & 46.28 & 34.12 & 78.43 & 17.13 & 17.82 \\
\midrule
\textsc{Oracle} & 94.52 & 85.40 & 77.58 & 65.10 & 47.84 & 86.58 & 20.60 & 22.59 \\
\bottomrule
\end{tabular}
}
\caption{Downstream results on KILT dev sets. Top---our best systems for Wikipedia and \sphere respectively. Bottom---a hybrid, oracle system which always chooses the better answer of the two above.}
\label{tab:reader_dev_upper}
\end{table*}

\subsection{Universal web retrieval}
\paragraph{Knowledge coverage of \sphere.}
Given that we use \sphere without any explicit alignment to the datasets we consider, the question of whether the corpus actually contains information necessary to solve the task at hand becomes of major importance.
First, we note that the popularity of the topic impacts how well represented it is on the web. This inadvertently leads to a limited knowledge coverage of rare topics when working with incomplete snapshots rather than an exhaustive index of the web. As a consequence, both slot-filling tasks suffer a large drop in performance when switching from Wikipedia to \sphere (see Section~\ref{sec:coverage} of the Appendix for more details).
Subsequently, we observe that on other tasks, \sphere is competitive with Wikipedia. The SOTA that the \fidbm architecture achieves on \trivia is our most salient result, outperforming our best model grounded in Wikipedia by over 6 points. \trivia can be considered one of the least Wikipedia-dependent of all KILT tasks---an encouraging evidence that web knowledge may be particularly useful in satisfying diverse information needs, especially those going beyond Wikipedia.
Finally, in Table~\ref{tab:reader_dev_upper}, we report results on KILT dev sets for the best systems using Wikipedia and \sphere respectively, and a hypothetical, hybrid, oracle system which is correct if either of them is correct. The oracle outperforms both baselines, suggesting that evidence provided by \sphere adds value on top of Wikipedia.

\paragraph{Wikipedia on the web.}
Based on a simple heuristic (details in Section~\ref{sec:wiki_on_web} of the Appendix), we estimate that over 5\% of \sphere passages are likely a copy from Wikipedia, with 47\% of Wikipedia passages having an equivalent in our web corpus.
Following this observation we note that all considered retrieval methods have a bias towards Wikipedia, surfacing a disproportionally high number of Wikipedia-based passages, with BM25 being the least biased. In the Appendix, we analyze the impact of Wikipedia passages on the \sphere downstream results further.

\paragraph{Sparse vs. dense models.}
The \aic gains we observe when moving from \dprmulti to \dprweb on \sphere correlate well with downstream performance. However, we don't see a similarly strong dependency between \dprweb and BM25 (Figures \ref{fig:ans_ent_a} and \ref{fig:ans_ent_c})---though the former often achieves better \aic scores, it lags downstream for all datasets but \nq.
To explain this, we ablate on the number of retrieved passages (see Figure~\ref{fig:ndoc_ablation} in the Appendix). The fewer contexts we consider, the smaller the BM25 advantage---if we use only the top one, the \dprweb-based model is better across the board, correlating well with \aic{@1}. This suggests that while \dprweb is able to find a good top document, the quality of the larger result set is worse---possibly because of false positives introduced when using \aic as retrieval supervision.
We investigate result set quality further by looking at the \aeic metric. Here, BM25 achieves the best results on all datasets except \nq (Figure~\ref{fig:ans_ent_d}), correlating better with downstream performance. We check how often the main entity is present in the input itself---\nq is an outlier in this regard, with the lowest fraction of datapoints containing the main entity (Table ~\ref{tab:entity_in_input}). It has been shown previously that BM25 is better at lexical exact-match on the salient spans in the query \cite{chen2021salient}. In our experiments BM25 can leverage this advantage---however, if the queries are more challenging in this regard as it is in the case with \nq, DPR becomes competitive.

\begin{table*}[t!]
\centering
\centering
\resizebox{\textwidth}{!}{    
    \fontsize{8.4}{10.1}\selectfont \setlength{\tabcolsep}{0.5em}
\begin{tabular}{ccccccccccc}
\toprule
Model  & \textbf{COPA} & \textbf{PIQA}  & \textbf{H-SWAG} & \textbf{CSQA} & \textbf{\textalpha NLI} & \textbf{NumS} & \textbf{WG} & \textbf{SocIQa} & \textbf{CosQA}   \\ 
\midrule
T5-large (no retrieval)             & 84.00          & 78.67          & 79.84          & 72.56           & 77.48          & 59.71          & 76.48          & 74.16          & 79.23      \\
Wikipedia (\fidbm) & 83.00          & 79.65          & 79.96          & \textbf{73.63}  & \textbf{77.94} & 62.30          & 76.72          & \textbf{74.36} & 78.83        \\ 
\sphere (\fidbm)  & \textbf{85.00} & \textbf{81.66} & \textbf{81.96} & \textbf{73.63}  & 77.74          & \textbf{66.70} & \textbf{76.80}  & 73.64  & \textbf{79.63}       \\

\bottomrule
\end{tabular}
}
\caption{Downstream accuracy on the dev set per common sense task.}
\label{tab:commonsense_results}
\end{table*}

\paragraph{Conclusions.}

Even though neural retrievers such as DPR beat BM25 by a large margin on Wikipedia, we haven't been able to apply them to \sphere with a similar success. It was suggested before that bi-encoders may be inherently not expressive enough for the purpose of large scale retrieval \cite{10.1162/tacl_a_00369}. Still, we do see avenues for improvement. \aic may be too weak of a signal for retrieval supervision, with \aeic emerging as a potential alternative. In addition, our DPR models display a bias towards Wikipedia-based results which could be mitigated by picking better positive samples for finetuning from the web. In line with previous research \cite{maillard-etal-2021-multi, oguz2021domainmatched} we also note that zero shot transfer of DPR models
doesn't yield good results, leaving the challenge of building universal, neural web retrievers open.

\subsection{Common Sense Tasks}
Rather than competing with the state of the art, which, for many common sense tasks, can be achieved with billion-parameter-scale, closed-book language models \cite{Lourie2021UNICORNOR}, we propose a proof-of-concept experimental setup. Our goal is to validate a hypothesis that knowledge augmentation with a general-purpose web index can positively impact the performance of an end-to-end system on common sense tasks. Table \ref{tab:commonsense_results} contains downstream results for Wikipedia and \sphere-augmented \fid models, as well as a comparable in size, closed-book, T5-large baseline.
We observe that retrieval brings consistent gains, with \sphere providing a clearly better improvement than Wikipedia on COPA, PIQA, HellaSWAG and NumerSense, indicating that it can serve as a broader source of common sense knowledge.
When investigating the passages retrieved from \sphere, we find that they rarely surface \emph{explicit} rules or generic statements expressing common sense knowledge. Rather, the retriever finds \emph{instances} of real world situations that serve to build an on-the-fly, common sense understanding of the problem at hand (see examples in Table \ref{tab:commonsense_examples} in the Appendix). This poses an interesting challenge to the reader which needs to infer general rules based on specific illustrations of their application - like in the CommonsenseQA example, where the model should predict that people like to have coffee in the office based on a description of a dream office with a coffee table in it.

\section{Related Work}

Most existing research into factual KI-NLP uses Wikipedia as the source of knowledge \cite{kwiatkowski-etal-2019-natural,joshi2017triviaqa,Thorne18Fever,yang2018hotpotqa,dinan2018wizard,petroni-etal-2021-kilt}. In this paper, we instead study our ability to solve KI-NLP tasks with web as the background corpus.
Previous works that operate on web \cite{joshi2017triviaqa,bajaj2018ms,talmor-berant-2018-web} typically rely on results from black-box search engines to create a corpus. A \ccnet snapshot has been considered as a knowledge source in dialog research by \citet{komeili2021internetaugmented}, where authors use it together with a Wikipedia snapshot. As far as we know, our work is the first to consider an uncurated snapshot of the web \emph{without} Wikipedia as a knowledge source for multiple KI-NLP tasks at once. Moreover, our scale is significantly larger than previously attempted (see Table \ref{tab:corpus_size}). There are other large scale resources that could be considered to tackle KI tasks, such as large collections of question-answer pairs~\cite{10.1162/tacl_a_00415,huber2021ccqa}, structured knowledge sources~\cite{berant-etal-2013-semantic,levy2017zero,elsahar2019t} or domain specific collections~\cite{tsatsaronis2015overview,saikh2021covidread}.
As for the common sense NLP tasks, while large pretrained models have achieved remarkable performance \cite{raffel2019exploring, brown2020language,Lourie2021UNICORNOR}, researchers have been seeking external repositories of common sense to boost performance further \cite{mitra2019additional, lin2021differentiable, xu2021human}. Existing common sense resources include both structured knowledge bases \cite{fellbaum2010wordnet, speer2012conceptnet, sap2019atomic} and natural language statements \cite{sumithra2020genericskb}. To the best of our knowledge, ours is the first work exploring common sense retrieval from a web corpus at this scale.

\section{Discussion and Future Work}
Harnessing the vast textual resources available online today through white-box retrieval may be the source of the next big break in NLP. In our current work, we propose to use a web snapshot as a universal, uncurated and unstructured knowledge source for multiple factual and common sense knowledge tasks at once.
We see encouraging results even in the experimental setup with a strong pro-Wikipedia bias, which suggests that \sphere is a competitive knowledge source with the potential of pushing the state of the art---especially for tasks with diverse information needs. At the same time, while remaining closer to the needs of real-world applications, our setup exposes limitations of existing retrievers, providing a challenging test bed for future innovations.
One of the key problems, which we aim to address in the future, regards the quality of retrieved information. Using Wikipedia as the knowledge source allows researchers to assume the high quality of the corpus documents. When transitioning to a web corpus, we no longer have the certainty that any document is good, truthful or unique, or that a certain \emph{gold document} containing all the necessary information even exists. Future work should focus on the ability of the models to assess the quality of the retrieved documents, handle duplicates, detect potential false claims and contradictions, prioritize more trustworthy sources and refrain from providing the answer if no sufficiently good evidence exists in the corpus.


\bibliography{anthology,acl2021,KILT}

\clearpage
\appendix

\section{Appendix}

\subsection{Knowledge coverage in \sphere}
\label{sec:coverage}

\begin{table}[ht]
\centering
\resizebox{\columnwidth}{!}{    
    \fontsize{7}{8}\selectfont \setlength{\tabcolsep}{0.5em}
    \begin{tabular}{R{3cm}cc}
    
    \toprule
    predicate & count & accuracy \\
    \midrule
    \multicolumn{3}{c}{\textbf{\zsre}}  \\
    \midrule
    
    mouth of the watercourse & 692 & 25.43 \\
    employer & 650 & 50.77 \\
    production company & 459 & 28.54 \\
    spouse & 357 & 41.46 \\
    from fictional universe & 340 & 16.18 \\
    crosses & 327 & 70.95 \\
    time of spacecraft launch & 295 & 33.9 \\
    drafted by & 287 & 68.64 \\
    date of official opening & 117 & 36.75 \\
    occupant & 112 & 4.46 \\
    
    \midrule
    \multicolumn{3}{c}{\textbf{\trex}} \\
    \midrule

    country & 617 & 88.17 \\ 
    located in the administrative territorial entity& 470 & 35.96 \\ 
    instance of & 464 & 68.32 \\ 
    country of citizenship & 344 & 85.17 \\ 
    taxon rank & 325 & 99.08 \\ 
    occupation & 311 & 74.28 \\ 
    sport & 278 & 84.89 \\ 
    place of birth & 215 & 37.21 \\ 
    performer & 147 & 27.89 \\ 
    parent taxon & 140 & 54.29 \\ 

    \bottomrule
    \end{tabular}
}
\caption{Top ten predicates in our slot-filling tasks, with per-predicate accuracy on the dev set.}
\label{tab:predicates}
\end{table}

\paragraph{Slot-fillig tasks.}
Slot-filling tasks suffer the biggest drop in downstream performance (see Table~\ref{tab:reader_test} in the main paper) when moving from Wikipedia to \sphere, which we investigate further by exploring their per-predicate accuracy (see Table~\ref{tab:predicates}). We observe a high variance in accuracy for the most common predicates, with those referring to more general concepts (e.g. \textit{crosses}, \textit{country}) scoring higher than more specific ones (e.g \textit{occupant}, \textit{performer}). We further note that all non-slot filling tasks incorporate a notion of input popularity in the data collection process. In \fever, claims were collected for \say{approximately 50,000 popular [Wikipedia] pages. These consisted of 5,000 from a Wikipedia \say{most accessed pages} list and the pages hyperlinked from them}, \nq contains aggregated, real-world search engine questions, in \hotpot, authors \say{manually curate 591 categories from the lists of popular pages by WikiProject} to source their questions from. Finally, \trivia questions were collected from trivia-related websites and further filtered to only include questions with high-quality search results. On the contrary, both \trex and \zsre triplets were sourced from unfiltered WikiData snapshots, and further sampled uniformly to match KILT size limits. This suggests the knowledge coverage on the web is not balanced, with popular topics receiving more representation than rare ones.

 \begin{table*}[ht]
\centering
\resizebox{\linewidth}{!}{    
    \fontsize{8.6}{10.5}
    \selectfont
    \setlength{\tabcolsep}{0.5em}

\begin{tabular}{R{.16\textwidth}p{.16\textwidth}p{.38\textwidth}p{.3\textwidth}}
    \toprule
    
    \textbf{Input} & \multicolumn{3}{p{.8\textwidth}}{\textit{After he had directed "Australia", it was reported that Baz Luhrmann's next project was a film based on which book by F Scott Fitzgerald?}}  \\
    \textbf{Gold Answer} & The Great Gatsby \\
    \midrule
    \multicolumn{2}{p{.34\textwidth}}{Top \sphere} & Top trivia & Top Wikipedia (Gold) \\
    \midrule
    \multicolumn{2}{p{.34\textwidth}}{
    {\fontfamily{cmss}\selectfont{\ldots Baz Luhrmann Adapting Fitzgerald's The Great Gatsby Next \textit{Although Australia sadly ended up as one of this year's biggest flops (currently at \$39 million), filmmaker Baz Luhrmann has already started work on his next project - F. Scott Fitzgerald's The Great Gatsby.} Nikki Finke of Deadline Hollywood confirms that this is Luhrmann's next gig and that he's already actively searching for a young actress to portray Daisy in the film. This isn't the first time Fitzgerald's book has been adapted - there was a 1974 film directed by Jack Clayton as well as a 1949 film directed by \ldots}}}
    &
    {\fontfamily{cmss}\selectfont{\ldots The publication of which book by Salman Rushdie led to threats on his life by Ayatollah Khomeini? On the last day of his life Bhagat Singh was reading a book about the Ideology of which revolutionary ? Who wrote the book "Life of Pi"? \textit{After he had directed "Australia", Baz Luhrmann's next project was a film based on which book by F Scott Fitzgerald?} Who advocated that a free market economy is more productive and more benefial to society, in his famous book ? Who is the author of the book titled "A Kingdom For His Love"? The publication of which book by Salman Rushdie led to threats on his life by Ayatollah Khomeini? posted Jan 17\ldots}}
    &
    {\fontfamily{cmss}\selectfont{\ldots On the screen he is best known for his Red Curtain Trilogy, comprising his romantic comedy film "Strictly Ballroom" (1992), the romantic tragedy "William Shakespeare's Romeo + Juliet" (1996), and "Moulin Rouge!" (2001). \textit{Following the trilogy, projects included "Australia" (2008), "The Great Gatsby" (2013),} and his television period drama "The Get Down" for Netflix. Additional projects include stage productions of Giacomo Puccini's\ldots}}
    \\
    
    \bottomrule
\end{tabular}

}
\caption{The \sphere corpus contains passages from trivia-devoted web pages featuring questions from the \trivia dataset, however, these passages typically don't provide any context information and often don't contain answers. Filtering them out doesn't significantly impact downstream performance.}
\label{tab:examples_quizz}
\end{table*}

\paragraph{TriviaQA.}
A \sphere-grounded \fid model achieves a SOTA performance, beating our best Wikipedia-based model by over 6 points. We note that \trivia is an outlier among other datasets and it can be considered one of the least Wikipedia-dependent of all KILT tasks. Questions and answers in \trivia were created independently by trivia enthusiasts and only distant supervision was applied to collect Wikipedia evidence. We test a hypothesis that the \sphere advantage over Wikipedia might result from the fact that it would contain trivia websites with questions from the dataset. We find this not to be the case though: filtering out passages which contain input questions verbatim from the result sets of respective samples does not meaningfully impact downstream performance (see Table~\ref{tab:examples_quizz} for more context).

\subsection{Wikipedia vs. the web}
\label{sec:wiki_on_web}

\paragraph{Wikipedia dissemination on the web.}
Excluding Wikipedia URLs from \sphere was an early design decision. However, Wikipedia text dissemination on the web goes beyond Wikipedia itself. We apply a simple \emph{ngram filtering} heuristic testing if a web passage has at least one 8-gram overlap with a Wikipedia passage to establish if it was based on Wikipedia (a method inspired by \citet{radford2019language}). We will refer to such a passage as \emph{wiki-based}. First, we note that as much as 5\% of passages in our web corpus are wiki-based, adding up to almost 46M passages in total while the original Wikipedia corpus contains only 22M passages. This surprisingly high number can be partly explained by how \sphere was constructed - the \textit{head} \ccnet tier we used contains the documents with the lowest perplexity under a Wikipedia-based language model, favoring the inclusion of wiki-based passages into the corpus. We further note that almost 47\% of the passages present in the KILT knowledge source inspire at least one web passage in \sphere, suggesting that big chunk of Wikipedia has been copied somewhere on the web.

\begin{table*}[t!]
\centering
\resizebox{0.7\textwidth}{!}{    
    \fontsize{8.4}{10.1}\selectfont \setlength{\tabcolsep}{0.5em}
\begin{tabular}{cccccccccc}
 \toprule
 & \textbf{\fever}  & \textbf{\trex} & \textbf{\zsre}  & \textbf{\nq} & \textbf{\hotpot} & \textbf{\trivia} & \textbf{\eli}  & \textbf{\wow} & \cellcolor{gray!15}  \textbf{Avg.} \\

\midrule
\multicolumn{10}{c}{\textbf{\sphere}} \\ 
\midrule
\dprmulti & 32 & 21 & 21 & 20 & 30 & 19 & 10 & 18 & \cellcolor{gray!15} 22.62 \\
\dprweb & 29 & 20 & 29 & 23 & 26 & 24 & 18 & 25 & \cellcolor{gray!15} 24.25 \\
BM25  & 15 & 9 & 10 & 8 & 17 & 16 & 2 & 4 & \cellcolor{gray!15} 12.12 \\
\midrule
\multicolumn{10}{c}{wiki-based passages in \sphere: 5.07\%} \\
\multicolumn{10}{c}{Wikipedia passages with an overlapping passage in \sphere: 46.9\%} \\
\bottomrule
\end{tabular}
}
\caption{Median number of wiki-based passages among the top-100 results retrieved from \sphere for respective datasets and models followed by Wikipedia-\sphere overlap statistics. We consider passages to be overlapping if they share an 8-gram. }
\label{tab:wiki_in_web}
\end{table*}
\begin{table*}[t!]
\centering
\resizebox{0.7\textwidth}{!}{    
    \fontsize{8.4}{10.1}\selectfont \setlength{\tabcolsep}{0.5em}
\begin{tabular}{ccccccccc}
 \toprule
Model  & \textbf{\fever}  & \textbf{\trex} & \textbf{\zsre}  & \textbf{\nq} & \textbf{\hotpot} & \textbf{\trivia} & \textbf{\eli}  & \textbf{\wow}   \\
    \cmidrule(lr){2-4} \cmidrule(lr){5-7} \cmidrule(lr){8-8} \cmidrule(lr){9-9} 
    & \multicolumn{3}{c}{Accuracy} & \multicolumn{3}{c}{Exact Match} & \multicolumn{1}{c}{RL} & \multicolumn{1}{c}{F1} \\
\midrule
\fidweb & -2.01 &  -1.47 &  -5.20 & -7.96 & -11.47 & -0.45 & +0.76 & -4.71 \\

\fidbm & -1.84 &  -3.57 &
-10.42 & -6.17 & -12.26 &
-0.96 & 
-2.50 & -8.04 \\

\bottomrule
\end{tabular}
}
\caption{The relative change in the downstream performance when moving from the default, URL-based Wikipedia filtering startegy (see retults in Table~\ref{tab:reader_test}) to a more aggressive \emph{ngram} filtering strategy, in percents.}


\label{tab:wiki_filtering}
\end{table*}

\paragraph{Wikipedia bias in retrieval.}
We then look at the median number of wiki-based passages retrieved from \sphere for respective datasets in Table~\ref{tab:wiki_in_web}. It turns out that all retrieval methods have a bias towards Wikipedia - the average median number of wiki-based results retrieved by the BM25 retriever is 12.1 and it increases sharply for the DPR-based methods, with 22.6 for \dprmulti and 24.2 for \dprweb. \dprmulti is a Wikipedia retriever so it is not surprising that it is biased towards wiki-based passages. However, it is unexpected that fine-tuning leads to a retriever yielding even more wiki-based results than the original. The analysis from the previous paragraph may shed some light here: we estimate that as many as 34\% of DPR-based and 22\% of the BM25-based web training samples include wiki-based passages, so the fine-tuning process will reinforce the Wikipedia bias present in the baseline retrievers.

\begin{table*}[t!]
\centering
\resizebox{\linewidth}{!}{    
    \fontsize{8.4}{10.1}\selectfont \setlength{\tabcolsep}{0.5em}
\begin{tabular}{cccccccccccccccc}
 
\toprule
 & \multicolumn{3}{c}{\cellcolor{gray!15}\textbf{\trex}} & \multicolumn{3}{c}{\textbf{\zsre}}  & \multicolumn{3}{c}{\cellcolor{gray!15}\textbf{\nq}} & \multicolumn{3}{c}{\textbf{\hotpot}} & \multicolumn{3}{c}{\cellcolor{gray!15}\textbf{\trivia}} \\
\midrule

$k$ & \cellcolor{gray!15}1 &\cellcolor{gray!15} 20 &\cellcolor{gray!15} 100 & 1 & 20 & 100 &\cellcolor{gray!15} 1 &\cellcolor{gray!15} 20 &\cellcolor{gray!15} 100 & 1 & 20 & 100 & \cellcolor{gray!15}1 & \cellcolor{gray!15}20 &\cellcolor{gray!15} 100 \\
\midrule

\multicolumn{16}{c}{\aic} \\
\midrule

\multicolumn{16}{c}{\textbf{Wikipedia}} \\
\midrule

\dprmulti &\cellcolor{gray!15}76.36 &\cellcolor{gray!15} 94.54 &\cellcolor{gray!15} 96.74 &57.87 &90.82 &96.08 &\cellcolor{gray!15} 56.47 &\cellcolor{gray!15} 88.09 &\cellcolor{gray!15} 93.76 &30.64 &64.70 &76.52 &\cellcolor{gray!15} 69.75 &\cellcolor{gray!15} 94.61 &\cellcolor{gray!15} 98.10 \\
\dprweb &\cellcolor{gray!15} 84.28 &\cellcolor{gray!15} 96.34 &\cellcolor{gray!15} 97.62 &75.54 &96.62 &98.74 &\cellcolor{gray!15} 58.48 &\cellcolor{gray!15} 90.27 &\cellcolor{gray!15} 95.31 &35.48 &70.23 &81.00 &\cellcolor{gray!15} 73.13 &\cellcolor{gray!15} 96.12 &\cellcolor{gray!15} 98.64 \\

\midrule
\multicolumn{16}{c}{\textbf{\sphere}} \\
\midrule
\dprmulti & \cellcolor{gray!15} 40.08 &\cellcolor{gray!15} 67.38 &\cellcolor{gray!15} 76.16 &12.03 &36.47 &52.15 &\cellcolor{gray!15} 37.72 &\cellcolor{gray!15} 74.52 &\cellcolor{gray!15} 84.00 &17.63 &44.80 &59.66 &\cellcolor{gray!15} 61.73 &\cellcolor{gray!15} 91.88 &\cellcolor{gray!15} 96.83 \\
\dprweb & \cellcolor{gray!15} 57.50 &\cellcolor{gray!15} 78.12 &\cellcolor{gray!15} 83.84 &28.44 &54.73 &66.14 &\cellcolor{gray!15} 53.61 &\cellcolor{gray!15} 83.12 &\cellcolor{gray!15} 89.25 &30.30 &59.77 &70.48 &\cellcolor{gray!15} 78.99 &\cellcolor{gray!15} 96.08 &\cellcolor{gray!15} 98.12 \\
BM25 & \cellcolor{gray!15} 42.32 &\cellcolor{gray!15} 70.66 &\cellcolor{gray!15} 78.92 &22.45 &55.99 &67.86 &\cellcolor{gray!15}27.21 &\cellcolor{gray!15} 68.56 &\cellcolor{gray!15} 80.86 &25.36 &54.91 &68.48 &\cellcolor{gray!15} 67.66 &\cellcolor{gray!15} 96.81 &\cellcolor{gray!15} 99.01 \\

\midrule
\multicolumn{16}{c}{\aeic} \\

\midrule
\multicolumn{16}{c}{\textbf{Wikipedia}} \\
\midrule
\dprmulti & \cellcolor{gray!15} 59.90 &\cellcolor{gray!15} 86.26 &\cellcolor{gray!15} 90.24 &49.54 &86.76 &92.32 &\cellcolor{gray!15} 31.65 &\cellcolor{gray!15} 64.75 &\cellcolor{gray!15} 72.08 &27.66 &57.13 &66.93 &\cellcolor{gray!15} 47.45 &\cellcolor{gray!15} 82.40 &\cellcolor{gray!15} 89.23 \\
\dprweb &\cellcolor{gray!15} 68.08 &\cellcolor{gray!15} 89.04 &\cellcolor{gray!15} 91.56 &70.22 &94.52 &96.83 &\cellcolor{gray!15} 34.65 &\cellcolor{gray!15} 66.69 &\cellcolor{gray!15} 74.41 &31.80 &60.62 &70.27 &\cellcolor{gray!15} 53.44 &\cellcolor{gray!15} 84.96 &\cellcolor{gray!15} 90.60 \\

\midrule
\multicolumn{16}{c}{\textbf{\sphere}} \\
\midrule
\dprmulti &\cellcolor{gray!15}  4.60 &\cellcolor{gray!15} 12.14 &\cellcolor{gray!15} 16.16 &6.61 &19.76 &26.58 &\cellcolor{gray!15} 16.53 &\cellcolor{gray!15} 45.61 &\cellcolor{gray!15} 57.17 &10.38 &29.02 &39.45 &\cellcolor{gray!15} 37.90 &\cellcolor{gray!15} 75.93 &\cellcolor{gray!15} 84.77 \\
\dprweb  &\cellcolor{gray!15} 10.88 &\cellcolor{gray!15} 23.18 &\cellcolor{gray!15} 27.26 &20.30 &39.18 &44.90 &\cellcolor{gray!15} 27.11 &\cellcolor{gray!15} 57.24 &\cellcolor{gray!15} 66.83 &18.34 &43.36 &53.30 &\cellcolor{gray!15} 57.08 &\cellcolor{gray!15} 85.41 &\cellcolor{gray!15} 90.05 \\
BM25  &\cellcolor{gray!15} 18.02 &\cellcolor{gray!15} 35.74 &\cellcolor{gray!15} 42.74 &19.82 &51.85 &61.98 &\cellcolor{gray!15} 16.88 &\cellcolor{gray!15} 48.18 &\cellcolor{gray!15} 60.77 &21.14 &46.02 &57.96 &\cellcolor{gray!15} 55.36 &\cellcolor{gray!15} 87.98 &\cellcolor{gray!15} 92.61 \\

\bottomrule
\end{tabular}
}
\caption{Dev set \aic@k (top) and \aeic{@k} (bottom) for Wikipedia and \sphere indices.}
\label{tab:ans_ent}
\end{table*}

\begin{figure*}[ht]

\begin{subfigure}{.18\textwidth}
  \centering
  \includegraphics[width=\linewidth]{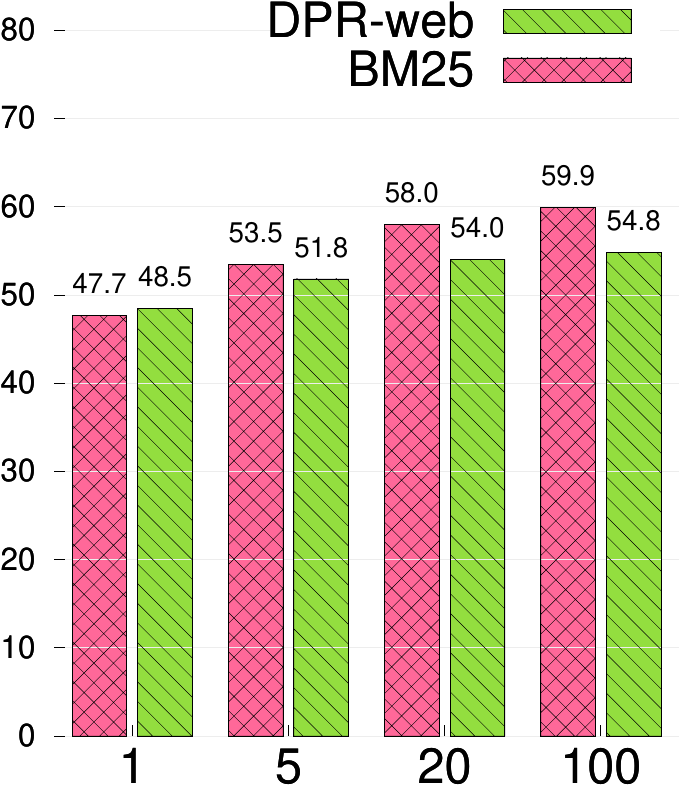}  
  \caption{T-REx.}
  \label{fig:trex_dvk}
\end{subfigure}\hspace{.01\textwidth}
\begin{subfigure}{.19\textwidth}
  \centering
  \includegraphics[width=\linewidth]{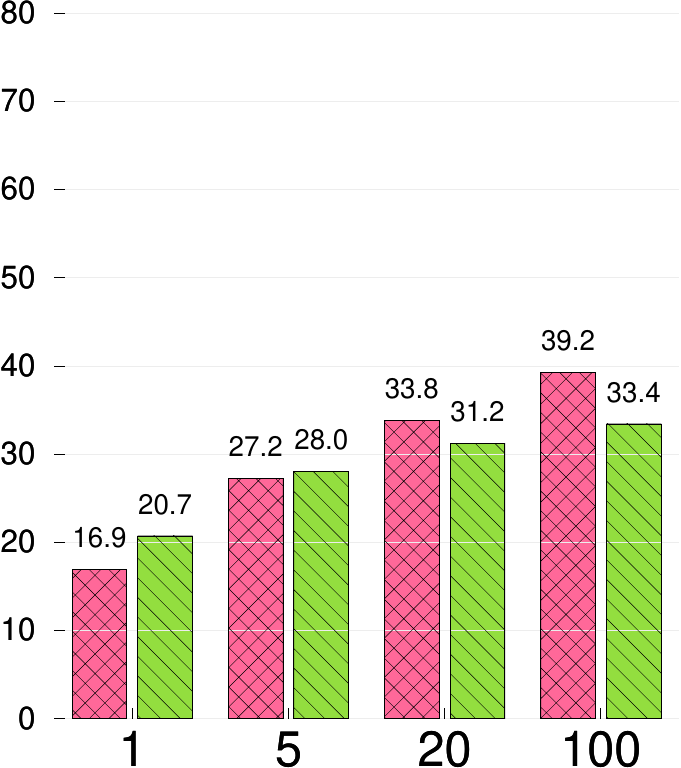}  
  \caption{zsRE.}
  \label{fig:zsre_dvk}
\end{subfigure}\hspace{.01\textwidth}
\begin{subfigure}{.19\textwidth}
  \centering
  \includegraphics[width=\linewidth]{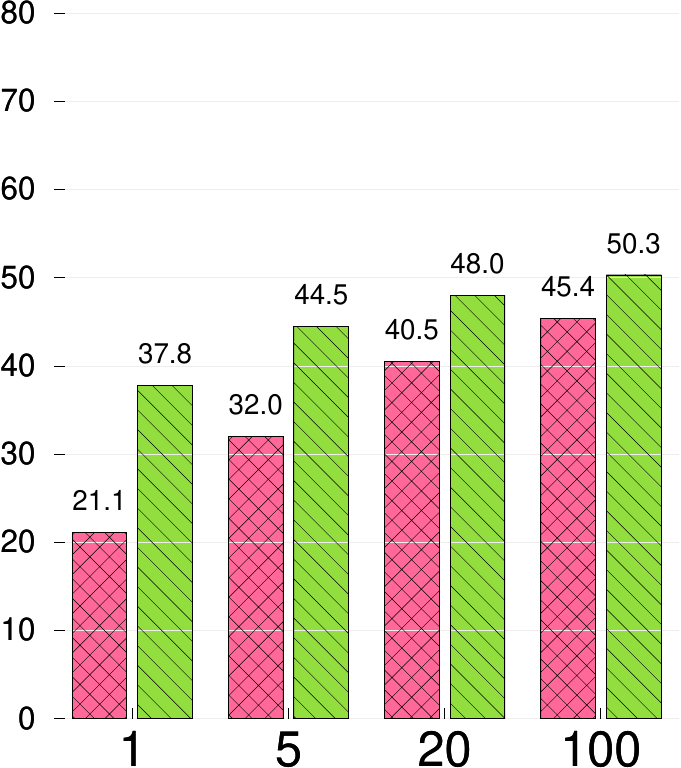}  
  \caption{NQ.}
  \label{fig:nq_dvk}
\end{subfigure}\hspace{.01\textwidth}
\begin{subfigure}{.19\textwidth}
  \centering
  \includegraphics[width=\linewidth]{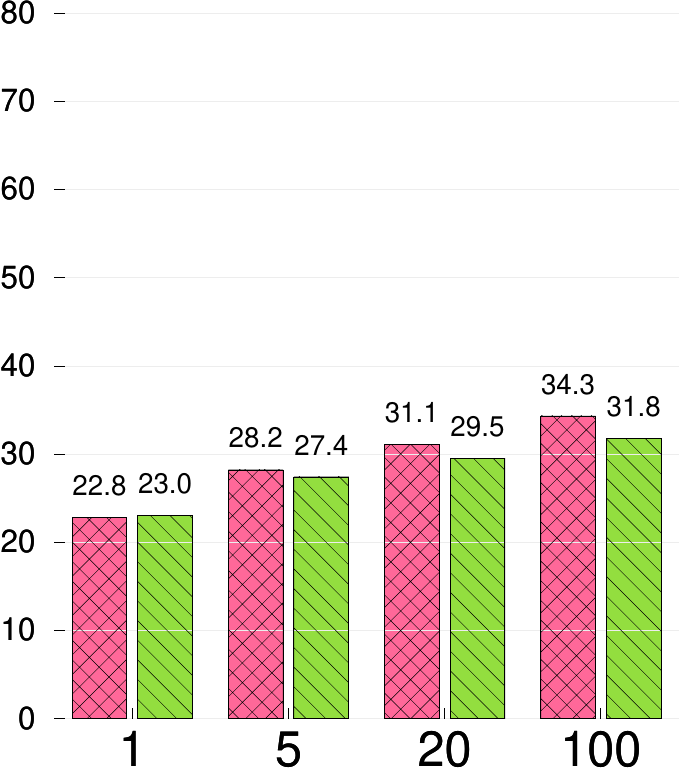}  
  \caption{HoPo.}
  \label{fig:hopo_dvk}
\end{subfigure}\hspace{.01\textwidth}
\begin{subfigure}{.19\textwidth}
  \centering
  \includegraphics[width=\linewidth]{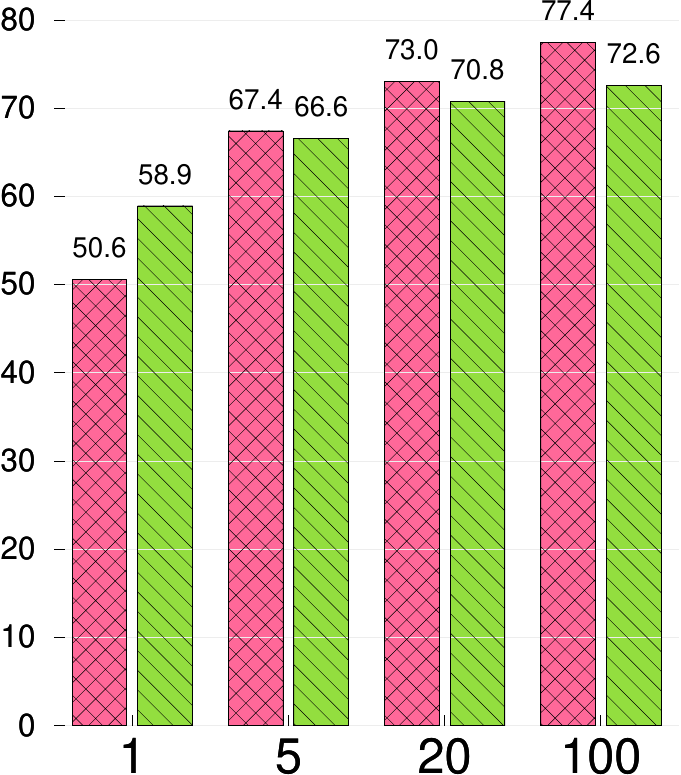}  
  \caption{TQA.}
  \label{fig:trivia_dvk}
\end{subfigure}
\caption{Dev set downstream evaluation results for \fid models with $k\in\{1,5,20, 100\}$ context passages. We plot accuracy for \trex and \zsre and exact match for \nq, \hotpot and \trivia.}\label{fig:ndoc_ablation}
\end{figure*}

\paragraph{Impact of Wikipedia on \sphere results.}
Finally, we seek to establish how much of the \sphere performance is thanks to the wiki-basesd passages contained in the corpus. In Table~\ref{tab:wiki_filtering}, we present the relative change in downstream results for \sphere-based models if we use the more aggressive \textit{ngram} filtering strategy. We generally see worse downstream results, however, the drop is not as dramatic as we would have expected. In particular on \trivia, even though suffering a small drop, the BM25-based architecture still obtains a SOTA performance with 77.46 points of exact match. These observations leave us optimistic about usefulness of the web as a knowledge source for KI-NLP tasks.

How to treat wiki-based passages when comparing web-based with pure Wikipedia-based solutions remains an open question. One may argue that aggressive filtering of wiki-based passages from the web would be the right course of action. At the same time, the quality of wiki-based copies is often poorer than the original and may degrade over time as Wikipedia gets updated. In a small subset of cases, we may also be facing a situation where the web inspires Wikipedia, not vice-versa. Ideally, we would want to aim for a retriever which would be able to recognize these situations and favor more reliable sources.

\begin{table*}[ht]
\centering
\resizebox{\linewidth}{!}{    
    \fontsize{8.6}{10.5}
    \selectfont
    \setlength{\tabcolsep}{0.5em}

\begin{tabular}{R{.07\textwidth}p{.29\textwidth}p{.29\textwidth}p{.35\textwidth}}
    \toprule
    & Gold & Wikipedia & \sphere \\
    \midrule
    \multicolumn{4}{c}{\textbf{\tqa}} \\
    \cmidrule(lr){2-4}
    \textbf{Input} & \multicolumn{3}{l}{\textit{What is the title of the film considered to be the debut of cartoon character Mickey Mouse?}}  \\
    \cmidrule(lr){2-4}
    \textbf{Answer} & Steamboat Willy, Timeless River, steamboat willie, Steamboat Willie, timeless river, steamboat willy, steam boat willie, Steam boat Willie & plane crazy & steamboat willie \\
    \cmidrule(lr){2-4}
    \textbf{Context}
    & 
    {\fontfamily{cmss}\selectfont{\ldots Mickey Mouse is a funny animal cartoon character and the mascot of The Walt Disney Company. He was created by Walt Disney and Ub Iwerks at the Walt Disney Studios in 1928. An anthropomorphic mouse who typically wears red shorts, large yellow shoes, and white gloves, Mickey is one of the world's most recognizable characters. Created as a replacement for a prior Disney character, Oswald the Lucky Rabbit, Mickey first appeared in the short "Plane Crazy", debuting publicly in the short film "Steamboat" \ldots}}
    & 
    {\fontfamily{cmss}\selectfont{\ldots "The Pointer" (1939) - "The Nifty Nineties" (1941) - "Lend a Paw" (1941) - "Symphony Hour" (1942) - "Squatter's Rights" (1946) - "Mickey and the Seal" (1948) - "The Simple Things" (1953) - "Mickey's Christmas Carol" (1983) - "Runaway Brain" (1995) - "Get a Horse!" (2013) Filmography Full- \ldots}}
    & 
    {\fontfamily{cmss}\selectfont{\ldots in the world. Mickey first was seen in a single test screening (Plane Crazy). Mickey officially debuted in the short film Steamboat Willie (1928), one of the first sound cartoons. He went on to appear in over 130 films, including The Band Concert (1935), Brave Little Tailor (1938), and Fantasia (1940). Mickey appeared primarily in short films, but also occasionally in feature-length films. Ten of Mickey's cartoons were nominated for the Academy Award for Best Animated Short Film, one of which, Lend a Paw, won the award in 1942. In 1978, Mickey became the first cartoon character to have a \ldots}} \\
    
    \toprule
    \multicolumn{4}{c}{\textbf{\fever}} \\
    \cmidrule(lr){2-4}
    \textbf{Input} & \multicolumn{3}{l}{\textit{Michelin Guides have been published for more than a decade.}}  \\
    \cmidrule(lr){2-4}
    \textbf{Answer} & SUPPORTS & REFUTES & SUPPORTS \\
    \cmidrule(lr){2-4}
    \textbf{Context}
    & 
    {\fontfamily{cmss}\selectfont{\ldots Michelin Guide Michelin Guides ( ) are a series of guide books published by the French tire company for more than a century. The term normally refers to the annually published Michelin "Red Guide", the oldest European hotel and restaurant reference guide, which awards up to three "Michelin stars" for excellence to a select few establishments. The acquisition or loss of a star can have dramatic effects on the success of a restaurant. Michelin also publishes a series of general guides to cities, regions, and countries, the \ldots}}
    & 
    {\fontfamily{cmss}\selectfont{\ldots Michelin Guide Michelin Guides ( ) are a series of guide books published by the French tire company for more than a century. The term normally refers to the annually published Michelin "Red Guide", the oldest European hotel and restaurant reference guide, which awards up to three "Michelin stars" for excellence to a select few establishments. The acquisition or loss of a star can have dramatic effects on the success of a restaurant. Michelin also publishes a series of general guides to cities, regions, and countries, the \ldots}}
    & 
    {\fontfamily{cmss}\selectfont{\ldots diners - or restaurant inspectors, as we better know them today - to visit and review restaurants anonymously. In 1926, the guide began to award stars for fine dining establishments, initially marking them only with a single star. Five years later, a hierarchy of zero, one, two, and three stars was introduced, and in 1936, the criteria for the starred rankings were published. During the rest of 20th century, thanks to its serious and unique approach, the MICHELIN Guides became best-sellers without equals: the guide now rates over 30,000 establishments in over 30 territories across three continents, and more than\ldots}} \\

    \bottomrule
\end{tabular}
}
\caption{Examples of datapoints in which our best \sphere-based architecture (\fidbm) outperforms our best Wikipedia-based architecture (\fidweb) for \tqa and \fever.}
\label{tab:examples}
\end{table*}

\begin{table*}[ht]
\centering
\resizebox{\linewidth}{!}{    
    \fontsize{8.6}{10.5}
    \selectfont
    \setlength{\tabcolsep}{0.5em}

\begin{tabular}{R{.07\textwidth}p{.46\textwidth}p{.46\textwidth}}
    \toprule
    & Wikipedia & \sphere \\
    \midrule
    \multicolumn{3}{c}{\textbf{COPA}} \\
    \cmidrule(lr){2-3}
    \textbf{Input} & \multicolumn{2}{p{.94\textwidth}}{\textit{Context: The couple travelled south for the winter.  Question: This happended because? \textbf{\color{Green}{Option 1: They were retired.}}  Option 2: They were separated.}}  \\
    \cmidrule(lr){2-3}
    \textbf{Answer} & 2 & 1 \\
    \cmidrule(lr){2-3}
    \textbf{Context}
    & 
    {\fontfamily{cmss}\selectfont{\ldots retirement, which put more strain on her marriage. In a speech commemorating her 25 years in parliament, she stated that her retirement was forced on her and that it should please the men of Britain. The couple began travelling separately and soon were living apart. Lord Astor also began moving toward left-wing politics in his last years, and that exacerbated their differences. However, the couple reconciled before his death on 30 September 1952. Lady Astor's public image suffered, as her ethnic and religious views were increasingly \ldots}}
    & 
    {\fontfamily{cmss}\selectfont{\ldots As we look back at how people retired, we would have seen that many people waited to travel until they were retired. They worked hard for 40 years and saved money. If they had enough they would travel. Often times that traveling meant buying an RV and going south for the winter and coming back near family in the summer. Or they would buy an RV and take off and travel west (or east) to see what they could for anywhere from a couple of months to a couple of years before coming back to a \ldots}} \\

    \midrule
    \multicolumn{3}{c}{\textbf{PIQA}} \\
    \cmidrule(lr){2-3}
    \textbf{Input} & \multicolumn{2}{p{.94\textwidth}}{\textit{Question: To get a stain out of clothes. Option 1: Wipe the stain with a rag and dish soap. \textbf{\color{Green}{Option 2: Use a tide pen to target the stain..}}}}  \\
    \cmidrule(lr){2-3}
    \textbf{Answer} & 1 & 2 \\
    \cmidrule(lr){2-3}
    \textbf{Context}
    & 
    {\fontfamily{cmss}\selectfont{\ldots dishwashing liquid to get rid of spidermites. Dish soap has also been used to deter aphids. In some instances, the dish soap may be toxic to plant leaves and cause them to "burn". Use of soap or dish detergent to help spread pesticide on plants is noted by University of Georgia extension service, but not recommended. - A solution of dishwashing liquid and water may be used to remove coffee, tea, olive oil, soda and fruit juice stains from fabrics. One dishwashing liquid brand has been used \ldots}}
    & 
    {\fontfamily{cmss}\selectfont{\ldots Mr. Clean Magic Eraser to wipe off marks. Or, if you don't have one on hand, sprinkle some baking soda on top of your dish soap to add an extra oomph of abrasion and clean as usual. Sponge the stain with cool water or soak the garment in cool water for 30 minutes. Use a GH Seal holder Tide To Go Stain Pen to remove as much of the stain as possible, then pretreat with a prewash stain remover, like Resolve Stain Stick and launder as usual. If your coffee had milk or cream in it, \ldots}} \\

    \midrule
    \multicolumn{3}{c}{\textbf{CommonsenseQA}} \\
    \cmidrule(lr){2-3}
    \textbf{Input} & \multicolumn{2}{p{.94\textwidth}}{\textit{Question: Where do people want to have a lot of coffee? Option 1: table. \textbf{\color{Green}{Option 2: office.}} Option 3: desk. Option. 4: kitchen. Option 5: ocean. }}  \\
    \cmidrule(lr){2-3}
    \textbf{Answer} & 4 & 2 \\
    \cmidrule(lr){2-3}
    \textbf{Context}
    & 
    {\fontfamily{cmss}\selectfont{\ldots Low caffeine coffee , please reference the table below adapted from USFDA estimates: Along with Arabica, several coffee producers are now offering options of low-caffeine coffee, which can provide a solution for those who do not want to make the switch to decaf. In nature, coffee grows with varying levels of caffeine. Given various environmental factors, certain beans will grow with more caffeine than others, thereby creating an opportunity to produce naturally low caffeine coffee. Western producers have not yet shown a desire to sort purchased bean lots by caffeine content \ldots}}
    & 
    {\fontfamily{cmss}\selectfont{\ldots I used to write in my kitchen but in recent years, I moved my office to our bedroom where I can shut the door on noise. I have allotted 1/3 of this room to my office with a comfy chair as well as a desk and shelves, file cabinet, etc. All I need for working. As for my dream office, I actually have a house plan that I keep at my desk (as a means of hope \& motivation.) It includes an office with lots of windows, space enough for a love seat, chair and coffee table, as well as a \ldots}} \\

    \bottomrule
\end{tabular}
}
\caption{Examples of commonsense tasks where the \sphere-based architecture (\fidbm) outperforms Wikipedia-based architecture (\fidbm). Gold answers in bold green.}
\label{tab:commonsense_examples}
\end{table*}

\end{document}